# Computational Analysis of Heart Rate Variability in Healthy Adults

María J. Lado[1], Arturo J. Méndez[1], Leandro Rodríguez-Liñares[1], Baltasar García Perez-Schofield[1], Pedro Cuesta-Morales[1], Brais Iglesias-Otero[2], Xosé A. Vila [1]

[1] Escola Superior de Enxeñaría Informática, IFCAE - Universidade de Vigo, Campus As Lagoas, Ourense, 32004, Spain.
[2] Universidad Politécnica de Madrid, Escuela Tecn Super Ingn Telecomunicac, Informat Proc & Telecommun Ctr, Speech Technol & Machine Learning Grp, Madrid, 28040, Spain

[1] mrpepa@uvigo.es, mrarthur@uvigo.es, leandro@uvigo.es, jbgarcia@uvigo.es, pcuesta@uvigo.es, brais.iglesias@upm.es, anton@uvigo.es

**Abstract** Heart Rate Variability (HRV) analysis is a key indicator of cardiac physiological state and aids in disease diagnosis. However, research on HRV parameters in healthy individuals remains limited, and no gold standard exists. This study evaluates HRV indices in 40 healthy adults (20 men, 20 women, aged 30–50) to improve HRV's clinical utility. Using computational methods for signal processing and data analysis, time, frequency, and nonlinear indices were analyzed to address five questions: (1) normality, (2) stability, (3) correlation, (4) reproducibility, and (5) consistency. Key findings: (1) Time-domain and nonlinear indices, particularly global and LF (low frequency), follow normal distributions, with gender differences noted. (2) Most indices are stable except HF (high frequency)-related ones. (3) High correlations in HF-related indices suggest redundancy, indicating only one is necessary in studies. (4) Comparisons with the Fantasia database revealed <10% error for most indices, except SD2 and SDNN in women (>15%). (5) Time-domain and nonlinear indices show low inter-study variability, while frequency-domain indices exhibit high variability, limiting cross-study comparisons. The selected indices—ApEn and IRRR (global variability), HRVi and SD2 (LF), and MADRR or rMSSD (HF)—are best suited for accurately representing HRV components and enhancing its clinical and research relevance.

## 1 Introduction

Human well-being is heavily influenced by health conditions, both subjective and objective. Among the subjective factors, elements like happiness, job satisfaction, and certain personality traits play a significant role in the health-related quality of life. However, these aspects can sometimes be challenging to assess objectively. Recent advances in computational methods and AI-based health assessment have shown that subjective health metrics can be complemented with intelligent data analysis to obtain more reliable insights [1].

Conversely, measuring explicit physiological parameters offers a concrete method for determining an individuals' health status [2]. In fact, health-related clinical research gains value from incorporating both patients' perspective on their health, and the analysis of objective measures.

A key physiological indicator of health conditions is the Heart Rate (HR), defined as the number of heartbeats per unit of time, typically ranging from 60-100 beats/min under resting conditions. A notable advantage of this measurement is that it can be easily obtained using basic non-invasive medical equipment.

The time between two consecutive heartbeats is known as the RR interval [3], measured as the interval between the R waves of each heartbeat. RR intervals vary from beat to beat, and the measure of variations in the time intervals between beats is termed Heart Rate Variability (HRV) [4]. HRV reflects the balance between the sympathetic and parasympathetic nervous systems, which act to increase and decrease the heart rate, respectively.

Among the various methods to analyse HRV, frequency analysis is a commonly used approach. The Total spectral power (TF) is computed as the sum of power across all frequencies. Moreover, examining the different frequency components can yield insightful information about the neurovegetative nervous system: HF (high frequency) components of HRV (0.15-0.5 Hz) are associated with the modulation of the vagal tone, whereas LF (low frequency) components (0.05-0.15 Hz) are influenced by both the sympathetic and parasympathetic systems [5].

An alternative method to studying HRV involves time-domain measures. These methods are typically straightforward to implement, as they are derived directly from the RR interval series [6]. Common indices include SDNN (standard deviation of RR intervals); IRRR (inter-quartile difference); pNN50 (proportion of adjacent RR intervals differing by more than 50 ms); SDSD (standard deviation of differences between adjacent RR intervals); rMSSD (square root of the mean of the squares of the differences between adjacent RR intervals); MADRR (median of the absolute differences between adjacent RR intervals); and HRVi (HRV triangular index, calculated as the integral of the intervals histogram divided by its maximum).

HR regulation is significantly influenced by non-linear phenomena, making non-linear statistical analysis a viable method for y HRV study [7]. A commonly used non-linear estimator is ApEn (approximate entropy), which measures the fluctuations, and thus predictability, in the RR time series. Parameters SD1 and SD2, derived from the Poincaré plot (an ellipse representing the correlation between adjacent RR intervals), are also widely used. SD1 and SD2 are calculated as the standard deviations perpendicular and parallel to the line of identity of the ellipse, respectively.

Regardless of the domain (time, frequency, or non-linear) each parameter extracts information from a specific band of the HR's signal spectrum. Based on this criterion, all these indices can be categorized into three groups [6]:

- Global indexes: TF, SDNN, IRRR, ApEn.
- LF-related indexes: LF, HRVi, SD2.
- HF-related indexes: HF, pNN50, SDSD, rMSSD, MADRR, SD1.

It has been shown that HRV analysis, providing valuable insights into physiological and pathological states, can be a beneficial tool for enhancing risk stratification [5] and detecting various health-related conditions. Accordingly, HRV spectral analysis has been effectively employed in the detection of sleep apnoea [8], diabetes [9], and hypertension [10]. More specifically, the integration of frequency and time-domain analysis has been utilized to identify acute cardiac dysfunction [11], and nonlinear statistics methods have been proven effective in detecting myocardial infarction [12].

HRV analysis has also been applied to studies involving healthy populations. It has been used to investigate topics such as the impact of exercise [13], lifestyle activities [14], and reactions to political campaign ads [15]. Additionally, research has linked specific HRV parameter values with age or gender [16,17], work stress [18], and coffee consumption [19].

Despite extensive research in HRV analysis, there remains a lack of consensus in the scientific community regarding reference values for healthy people under normal conditions. No official guidelines for standard values and methodologies have been established, and only a few studies focus on analysing commercially available HRV measurement tools [20], investigating the effects of factors like gender or body posture on HRV analysis [21], or determining normality values for healthy population [22, 23]. For instance, the study by Sammito and Böckelmann [24] provided reference values from short segments (less than 5 min) of 24 h ECG (electrocardiogram) recordings. This approach was criticized by Bauer et al. [25], leading to the development of a new methodology and subsequent publication of substantially different values for HRV index values by the original authors [26]. Other research in the field aims to elucidate variations in HRV values within the general population [27], among children [28], or in individuals who have modified their physical activity [29].

When planning an experiment to study HRV, several critical decisions must be made regarding which software tools to use, the parameters to estimate, and all aspects related to data collection [30-32]. A research study found in the literature addresses some of these methodological issues [33]. Similar computational strategies have been applied in related studies, for example, in the automated analysis of cardiac signals using neural networks [34], highlighting the value of combining rigorous methodology with advanced computational tools.

In [33], authors concluded that collecting data in the morning with subjects seated is optimal, and it is advisable to discard the first minute of the recording and filter the data to minimize artifacts and noise, if possible. However, a significant limitation to this research was the small sample size, consisting of only 4 men and 2 women.

In this work, we present an extensive computational study of heart rate variability (HRV) in a healthy population under basal conditions. HR data were digitally recorded and processed using computational methods following the methodology provided in [33], and the reproducibility of the experiment was assessed using publicly available databases. Furthermore, we performed a computational comparison with results from previously published studies to evaluate consistency and validate our approach.

# 2 Materials and Methods

## 2.1 Database and acquisition protocol

Forty healthy individuals aged 30-50 years (20 men and 20 women) participated in our study. None were on medication, and none of the women were pregnant. For data protection and privacy, each participant was assigned a confidential code to label their data. No personal information was collected except for age and gender. Ethical approval was granted by the Research Committee of the corresponding University, and the study was conducted in accordance with the ethical principles of the Declaration of Helsinki.

The protocol for database acquisition followed the guidelines and recommendations in a previous study [33]: one recording per subject was obtained, and RR signals were recorded under identical conditions for all participants (in the morning, seated at rest in a silent, isolated room). To minimize involuntary movements, the subjects were viewing a collection of images during the acquisition.

The device used for signal acquisition was a Polar® H10 chest strap [35], a non-invasive sensor equipped with Bluetooth and two electrodes for heartbeat detection. Each participant wore the strap while seated in front of a laptop, beginning to view a set of images, during which time the corresponding RR signal sequence intervals were transmitted to the laptop via Bluetooth.

Signals were captured using the open-source tool gVARVI (graphic heart rate Variability Analysis in response to audioVisual stImuli) [36]. Each recording lasted 4 minutes. In line with the recommendations from [33], the first minute of each recording was excluded.

For all the calculations related to post-processing and analysing the datasets, the RHRV software package [6] for the R programming language [37] was used.

## 2.2 Database and acquisition protocol

Post-processing the database involved the following steps:

1. Acquisition of the instantaneous HR signal: the software tool gVARVI was used to obtain a 4-minute record from each subject, each containing a sequence of RR intervals in milliseconds. As mentioned earlier, the first 60 seconds of each recording were excluded. Subsequently, the RR intervals were inverted and scaled to convert the RR sequences into heart rate signals in beats/s.
2. Filtering: the instantaneous HR signal can be affected with artefacts, resulting from interferences, the subject's movements, or miss-detected heartbeats. Since the quality of signals can impact HRV indices, an automatic filtering process was applied to eliminate these artefacts. The filtering algorithm employs two criteria: (i) it removes all HR values that deviate by more than one adaptive threshold from both the preceding and the following values, as well as from the average of the last beats, and (ii) it discards points that fall outside acceptable physiological ranges [38]. Following this automatic filtering, all records were manually inspected to remove any potential remaining artefacts using a graphical software tool.

3. HR interpolation: until this stage, HR signals are non-uniformly sampled, which is unsuitable for obtaining frequency HRV indices using the Fourier transform. Therefore, cubic spline interpolation with a sampling frequency of 4 Hz was employed to achieve uniformly spaced signals.
4. HRV estimation: using both instantaneous and interpolated HR signals, HRV indices were calculated in frequency, time, and non-linear domains:
   a) Frequency-domain analysis: spectrograms of the interpolated HR records were generated using the Short-Time Fourier Transform (STFT) on the detrended signals, employing the Welch approximation [39]. This was done using 1-minute-long Hamming windows with 1 second shift. The mean power value in the LF and HF frequency bands were calculated, as well as the total power (TF) in all bands [33].
   b) Time-domain analysis: the following indices were computed from the non-interpolated RR series in milliseconds: SDNN, IRRR, pNN50, SDSD, rMSSD, MADRR and HRVi.
   c) Nonlinear analysis: ApEn, SD1, and SD2 indices were computed from the non-interpolated RR time series in milliseconds.

## 2.3 Analysis

Our aim was to analyse the behaviour of HRV indexes in a healthy population and compare our findings with those reported in existing literature. This objective guided us to examine a set of five aspects, each addressing a specific question related to HRV analysis:

1. Normality (are HRV indices normally distributed?): the outcome of normality tests is influenced by the specific conditions of the problem and the distribution of the data. Therefore, it is not feasible to identify an optimal test for assessing normality. To ascertain whether indices are normally distributed, we applied four normality tests suited for small samples [40]: Shapiro-Wilk, Pearson chi-square, Kolmogorov-Smirnov (with the correction proposed by Lilliefors), and Anderson-Darling.
2. Stability (which HRV indices are more stable across subjects?): different HRV parameters exhibit varying degrees of reliability and robustness [33]. The Coefficient of Variation (CV), also known as the relative standard deviation, is defined as the ratio between standard deviation and the mean value and serves as a reliable indicator of stability. A low CV value indicates that the values are closely clustered around the mean, suggesting that the parameter under study is stable.
3. Correlation (are HRV indexes correlated?): if two parameters exhibit high correlation, it implies that one of them might be redundant. We investigated the correlation among estimators, dividing them into the following groups: global, LF-related, and HF-related indices. Spearman's correlation coefficient [41] was used since it does not require a linear relationship between the variables under study.
4. Reproducibility (are values of HRV indexes consistent?): the values obtained in our study were compared with those from the public Fantasia Database [42]. This database, available on the PhysioNet platform [43], consists of 40 ECG recordings (each 120 minutes long) from 20 young (21-34 years old) and 20 elderly (68-85 years old) rigorously screened healthy subjects (20 men, 20 women). The ECGs were recorded while the subjects were in a resting state and in sinus rhythm, as they watched the Disney film “Fantasia” [44] to help maintain wakefulness. The signals were digitized at 250 Hz, and heartbeats were annotated using an automated algorithm, then verified through visual inspection.
5. Consistency with other researchers' findings (are our values consistent with values obtained by other researchers?): numerous studies in the literature report HRV index values derived from healthy subjects. For comparability, we selected studies that met the following criteria: (i) they included a minimum of 30 healthy subjects, (ii) all subjects were at least 18 years old, and (iii) the records were at least 5 minutes in length.

# 3 Results

## 3.1 Normality tests

Normality was assessed for all obtained parameters, with subjects categorized by gender to account for gender differences in HRV values. The results are presented in table 1, where indices that adhere to a normal distribution in both genders in at least two tests (P value > 0.01) are highlighted.

Global indices follow a normal distribution in both genders, except for the TF parameter, which is normal only for men as per the Pearson test.

Among LF-related indices, SD2 consistently exhibits a normal distribution. HRVi demonstrates normality in nearly all tests, with the exceptions being Pearson and Anderson-Darling tests for women.

A non-normal distribution is more prevalent in HF-related indices. The most consistent is MADDR, showing normality in all but two tests for women. The remaining time-domain and non-linear parameters exhibit normality in some instances, but not sufficiently to be deemed as normally distributed.

In summary, (i) frequency domain parameters are not normally distributed, (ii) global and LF-related indices are normal, except for frequency domain ones, and (iii) MADDR is the only HF-related index that consistently follows a normal distribution.

Table 1: Normality of HRV parameters: emphasized rows are for indices that show normality (P > 0.01) for both genders in at least two tests. Each cell contains two values M/W: the first (M) is for men and the second one (W) corresponds to women.

| | | Shapiro-Wilk | Pearson | Lilliefors | Anderson-Darling |
|---|---|---|---|---|---|
| Global | ***SDNN*** | ***0.443/0.002*** | ***0.791/0.159*** | ***0.744/0.060*** | ***0.483/0.019*** |
| | ***IRRR*** | ***0.252/0.113*** | ***0.267/0.342*** | ***0.613/0.360*** | ***0.358/0.250*** |
| | TF | $3.7\text{x}10^{-4}/8.5\text{x}10^{-6}$ | 0.069/0.001 | $0.002/3.2\text{x}10^{-5}$ | $4.4\text{e-}04/8.2\text{x}10^{-7}$ |
| | ***ApEn*** | ***0.026/0.726*** | ***0.069/0.207*** | ***0.059/0.509*** | ***0.025/0.582*** |
| LF | LF | $8.6\text{x}10^{-6}/1.2\text{x}10^{-6}$ | $0.0005/4.1\text{x}10^{-6}$ | $6.9\text{x}10^{-7}/5.0\text{x}10^{-8}$ | $1.1\text{x}10^{-7}/1.7\text{x}10^{-9}$ |
| | ***HRVi*** | ***0.840/0.022*** | ***0.910/0.003*** | ***0.715/0.029*** | ***0.693/0.009*** |
| | ***SD2*** | ***0.745/0.012*** | ***0.434/0.267*** | ***0.802/0.131*** | ***0.816/0.077*** |
| HF | HF | $7.6\text{x}10^{-6}/9.9\text{x}10^{-6}$ | 0.003/0.003 | $3.1\text{x}10^{-6}/0.002$ | $1.9\text{x}10^{-7}/1.1\text{x}10^{-6}$ |
| | pNN50 | 0.003/0.015 | 0.004/0.0004 | 0.049/0.092 | 0.002/0.018 |
| | SDSD | $0.004/1.9\text{x}10^{-5}$ | 0.207/0.029 | 0.132/0.005 | $0.009/6.9\text{x}10^{-5}$ |
| | ***MADRR*** | ***0.002/0.274*** | ***0.016/0.663*** | ***0.049/0.545*** | ***0.002/0.484*** |
| | rMSSD | $0.004/1.9\text{x}10^{-5}$ | 0.207/0.029 | 0.132/0.005 | $0.009/7.0\text{x}10^{-5}$ |
| | SD1 | $0.004/1.9\text{x}10^{-5}$ | 0.207/0.029 | 0.132/0.005 | $0.009/6.9\text{x}10^{-5}$ |

## 3.2 Stability tests

In this section, we focused on indices that exhibit a normal distribution (emphasized in table 1), as well as HF-related indices demonstrating normal behaviour in at least one test for both genders, given their frequent use in the literature. The Coefficient of Variation (CV) was calculated for each set. CV is valuable for measuring dispersion in a data set, as it assesses the stability of the HRV indices around their mean values.

Table 2 lists CV values for the specified HRV indices. For global and LF-related indices, all exhibited values below 0.5. Among HF-related indices, MADRR was the most stable, with a CV of 0.78 for men and 0.33 for women. Other indices had CV values around 0.7, except for pNN50, which was the least stable parameter for both genders.

Table 2: Coefficient of variation for normally distributed HRV indices (M/W = men/women).

| | M | W |
|---|---|---|
| SDNN | 0.49 | 0.47 |
| IRRR | 0.48 | 0.38 |
| ApEn | 0.11 | 0.10 |
| **Global indexes** | | |

| | M | W |
|---|---|---|
| HRVi | 0.32 | 0.27 |
| SD2 | 0.44 | 0.46 |
| **LF-related** | | |

| | M | W |
|---|---|---|
| pNN50 | 1.14 | 0.94 |
| SDSD | 0.77 | 0.65 |
| MADRR | 0.78 | 0.33 |
| rMSSD | 0.77 | 0.65 |
| SD1 | 0.77 | 0.65 |
| **HF-related** | | |

## 3.3 Correlation tests

The next step involved analysing the correlation between HRV indices. Correlations were assessed independently within global, LF-related, and HF-related groups. Table 3 shows values of the Spearman's rank correlation coefficients (a benefit of this coefficient is its non-reliance on the linearity between the variables) within these index groups. A high correlation value between two indices suggests they measure similar characteristics, rendering one redundant. Conversely, a low correlation value indicates they represent different aspects of the cardiovascular control system and should both be analysed.

Table 3: Correlation coefficients among HRV indices (M/W = men/women)

| | | IRRR | ApEn |
|---|---|---|---|
| SDNN | M | 0.96 | -0.08 |
| | W | 0.71 | -0.28 |
| IRRR | M | - | 0.04 |
| | W | - | -0.19 |
| **Global indexes** | | | |

| | | HRVi |
|---|---|---|
| SD2 | M | 0.87 |
| | W | 0.53 |
| **LF-related** | | |

| | | SDSD | MADRR | rMSSD | SD1 |
|---|---|---|---|---|---|
| pNN50 | M | 0.97 | 0.95 | 0.97 | 0.97 |
| | W | 0.91 | 0.92 | 0.97 | 0.91 |
| SDSD | M | - | 0.98 | 1 | 1 |
| | W | - | 0.84 | 1 | 1 |
| MADDR | M | - | - | 0.98 | 0.98 |
| | W | - | - | 0.84 | 0.84 |
| rMSSD | M | - | - | - | 1 |
| | W | - | - | - | 1 |
| **HF-related** | | | | | |

From the analysis of table 3, conclusions regarding necessary and redundant indices can be drawn:

- Global indices: ApEn is not correlated with others and should be included in HRV analyses. While SDNN and IRRR show moderate correlation for women, only one may be necessary.

- LF-related indices: HRVi and SD2 demonstrate moderate correlation for women, indicating that they are not entirely interchangeable. Ideally, both should be analysed.
- HF-related indices: all parameter pairs exhibit a high correlation coefficient (greater than 0.84). Therefore, it is reasonable to assume that all indices convey the same information, and only one is needed in HRV studies.

### 3.4 Reproducibility tests

Lack of reference values often impedes the clinical application of HRV parameters and complicates the comparison of results across different studies. To address this, we compared results from our database with those from the Fantasia Database [42].

We selected only recordings from young individuals in the Fantasia Database. After a visual inspection to identify, clean, artifact-free HR signals, 12 acceptable recordings remained (7 men and 5 women). Following recommendations given in [33], we discarded the first minutes from each two-hour signal and then post-processed and analysed the subsequent 3-minute segments (as described in the section Database and acquisition protocol).

For both databases, we calculated mean values of HRV parameters for men and women. For each index, an error rate was computed as

$$\%error = 2 \times \frac{|val_{this} - val_{fant}|}{val_{this} + val_{fant}} \times 100$$

where, for each HRV index, valthis and valfant are the values obtained using our database and the Fantasia database, respectively.

Error rates are shown in Figure 1. For men, differences were consistently below 8%, with particularly low values (below 3%) for SDNN, HRVi, SD2, MADRR, and rMSSD. For women, differences were less than 10% except for SDNN and SD2, with only ApEn and rMSSD indices yielded an error rate lower than 5%.

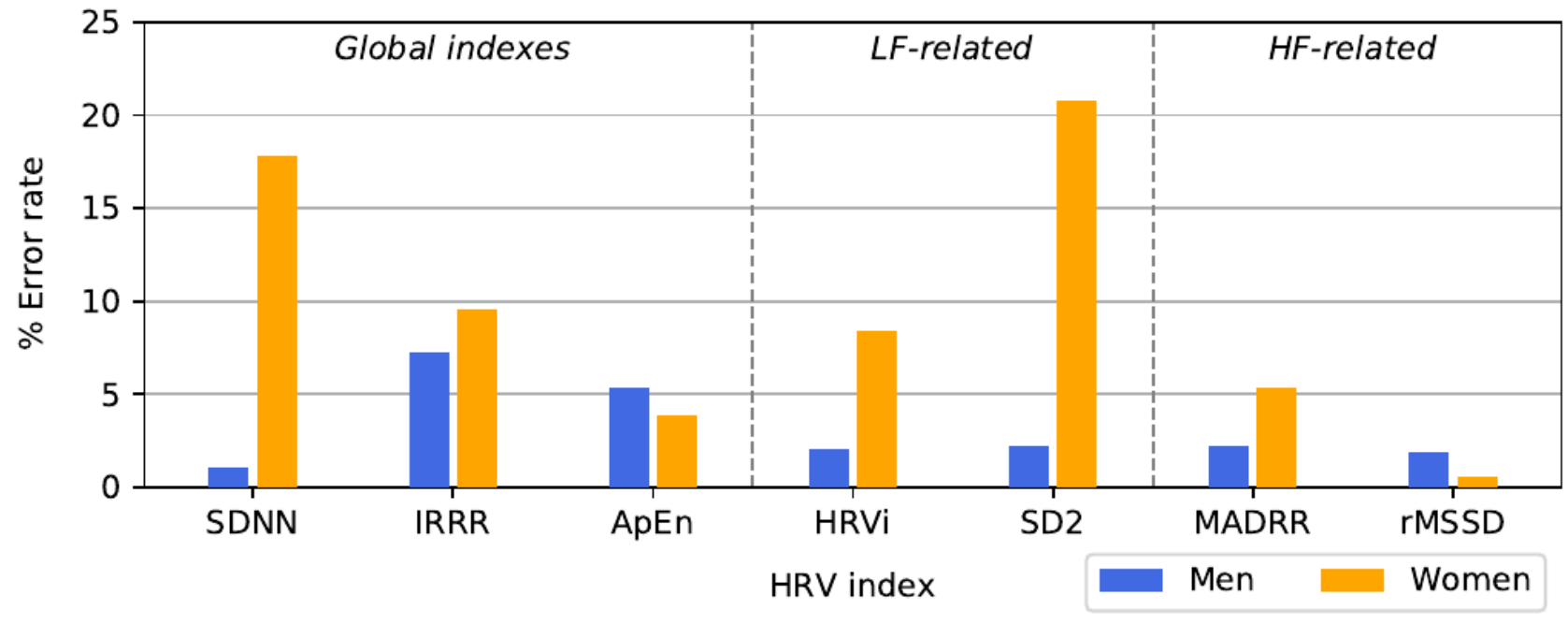

Figure 1. Differences in HRV indices from our work and from the Fantasia database.

### 3.5 Consistency tests

Comparing results with other published studies is crucial for evaluating the utility of HRV indices. We compared our results with studies that met the criteria outlined in the Analysis Section (at least 30 healthy subjects aged 18 or older and record lengths of at least 5 minutes). It is important to note that this comparison is not exhaustive: its primary purpose is to highlight differences and similarities, rather than to extensively analyse other studies.

Figure 2 graphically represents mean and standard deviation values of time-domain and non-linear HRV indices reported in five qualifying papers, compared with results from our database and the Fantasia Database. The numerical results for mean and coefficient of variation are included in table 4.

Table 4: Mean and CV of HRV indices (M/W = men/women) of HRV parameters reported in five papers meeting the requirements vs. results obtained with our database and with Fantasia.

| | | Global indexes | | LF-related | HR-related | |
|---|---|---|---|---|---|---|
| | | SDNN (ms) | ApEn | LF ($ms^2$) | HF ($ms^2$) | rMSSD (ms) |
| Melanson [29] | M | 51 (0.35) | - | 312 (0.58) | 919 (1.00) | 42 (0.52) |
| Pikkujämsä [22] | MW | 49 (0.43) | 1.14 (0.16) | 654 (1.00) | 383 (1.51) | - |
| | M | 52 (0.46) | 1.14 (0.14) | 747 (1.08) | 416 (1.78) | - |
| | W | 46 (0.35) | 1.14 (0.17) | 566 (0.78) | 351 (1.07) | - |
| Sandercock [20] | MW | 53 (0.39) | - | 918 (1.07) | 1194 (1.20) | 46 (0.54) |
| Nunan {23] | MW | 50 (0.10) | - | 519 (0.56) | 657 (1.18) | 42 (0.37) |
| Rajkumari [16] | M | 39 (0.36) | - | 674 (0.97) | 475 (0.85) | 27 (0.40) |
| | W | 41 (0.35) | - | 739 (0.78) | 491 (0.66) | 29 (0.45) |
| Our study | MW | 54 (0.47) | 0.99 (0.11) | 616 (1.41) | 347 (1.35) | 40 (0.71) |
| | M | 53 (0.49) | 0.96 (0.11) | 620 (1.38) | 364 (1.45) | 42 (0.77) |
| | W | 55 (0.47) | 1.01 (0.10) | 611 (1.47) | 329 (1.25) | 39 (0.65) |
| Fantasia DB | MW | 50 (0.35) | 0.96 (0.12) | 485 (0.62) | 236 (1.01) | 40 (0.42) |
| | M | 53 (0.21) | 0.91 (0.10) | 572 (0.47) | 228 (0.88) | 41 (0.32) |
| | W | 46 (0.54) | 1.05 (0.07) | 362 (0.92) | 247 (1.25) | 39 (0.60) |

Spectral parameters exhibit significant variation: the ranges for LF and HF are 312-918 and 228-1194, respectively, with most cases showing a CV greater than 1. This indicates that the utility of these indices is primarily limited to comparing different segments of a single recording or signals obtained and processed under identical conditions.

The time-domain indices, such as SDNN and rMSSD, display considerably less variation compared to frequency-domain ones (LF and HF). Most studies report similar SDNN values ranging from 49 to 54, with the only exception being study [16], and a CV consistently below 0.55. A similar pattern is observed with rMSSD, where studies report values between 39 and 46, again with the exception of [16]. In this case, CVs are 0.6 or lower except for our work, which shows 0.77 for men. The non-linear index ApEn was estimated in only one other study besides ours, yielding the most stable results with a CV below 0.2 in both cases.

Gender was considered separately in two studies besides ours. The results are not conclusive: one study [22] found HRV index values to be higher in men, while another [16] found them higher in women except LF. In our study, all indices are higher in men except ApEn.

# 4 Discussion

This paper presents a study on healthy subjects to enhance the clinical utility of HRV parameters. We collected HR data from healthy individuals and conducted HRV analysis focusing on the normality, stability, and reproducibility of results. Our acquisition protocol was taken from a previous study [33], specifically regarding body posture, time of day, and quality and duration of recordings.

All time domain and non-linear global indices exhibit a Gaussian distribution (table 1) and have CV values below 0.5 (table 2). ApEn emerged as the most stable global index, with CV values around 0.1 for both genders. Additionally, ApEn is not correlated with SDNN and IRRR, although these two indices are correlated with each other, suggesting only one might be necessary (table 3). Between SDNN and IRRR, despite SDNN’s popularity, our tests highlight two advantages of IRRR:

(i) it adheres to a Gaussian distribution in all tests for both genders (SDNN is only normal for men, according to Shapiro-Wilk test) and, (ii) it exhibits similar behaviour with both our database and the Fantasia Database for both genders (Figure 1).

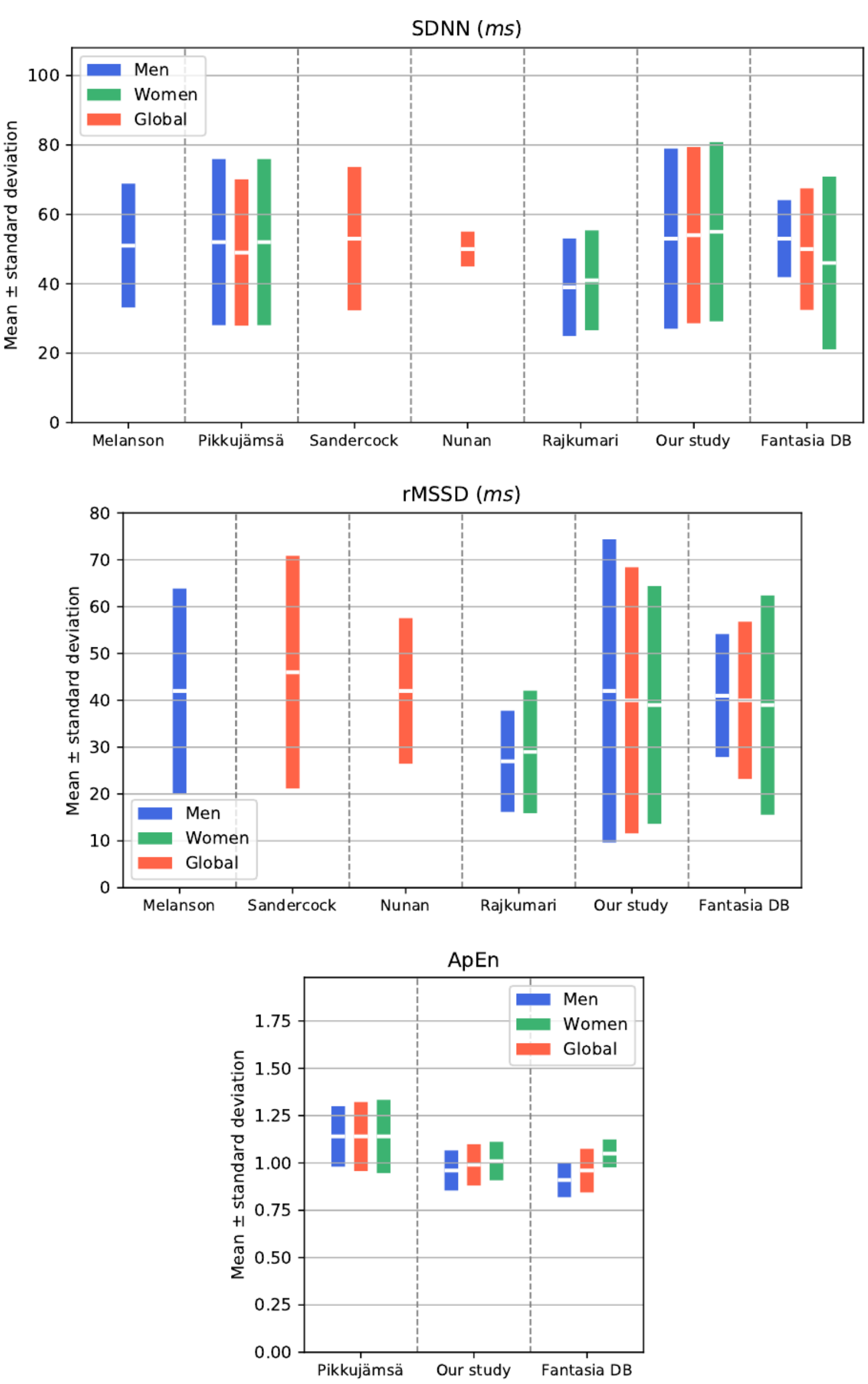

Figure 2. Mean ± standard deviation of time-domain and non-linear HRV parameters reported in five papers meeting requirements (Melanson [29], Pikkujämsä [22], Sandercock [20], Nunan [23], Rajkumari [16]) vs. results obtained with our database and Fantasia DB. The vertical range of the plots extends from zero to double the global mean obtained in our study, facilitating a clear view of the indices' dispersion.

In LF-related indexes, SD2 and HRVi exhibit some normal behaviour, unlike LF, which fails all normality tests (table 1). They are highly correlated for men but less so for women (able 3). Hence, if possible, both should be analysed.

Normality tests indicate that time-domain and non-linear HF-related parameters behave differently for women and men, corroborating previous reports in the literature [17,45,46]. In stability tests, MADRR emerged as a particularly stable index, especially for women (CV value of 0.33). Table 2 shows similar trends for other HF-related indices, further indicating gender-based differences. Correlation tests revealed that all time domain and non-linear parameters essentially convey the same information (table 3); therefore, one would suffice for studying high frequency variations of HRV. For these reasons, while MADDR is a suitable choice, rMSSD is more commonly used by researchers, and our tests indicated smaller differences across databases for rMSSD.

In summary, our recommendation is to use ApEn and, preferably IRRR over SDNN for quantifying global variability, HRVi and SD2 for LF, and MADRR or rMSSD for HF estimators.

All our results (from both our database and the Fantasia Database) were derived from young, healthy subjects. Therefore, extrapolating these findings to populations with different characteristics should be approached cautiously. Bearing this in mind, we calculated HRV indices for both young and elderly subjects in the Fantasia database. Table 5 shows these values (not separated by gender), and it is evident that they differ significantly (t-student test showed statistically significant differences, i.e., P values below 0.05). This suggests that elderly subjects exhibit lower HRV than young individuals, in agreement with existing literature.

Table 5: HRV indices and t-test results (P values) for young and elderly subjects of the Fantasia Database.

| | **Global indexes** | | | **LF-related** | | **HF-related** | |
|---|---|---|---|---|---|---|---|
| | **ApEn** | **IRRR** | **SDNN** | **HRVi** | **SD2** | **MADDR** | **rMSSD** |
| Young | 0.96 | 66.99 | 50.24 | 10.97 | 64.78 | 26.08 | 40.14 |
| Elderly | 0.84 | 44.12 | 33.52 | 7.72 | 45.34 | 12.35 | 18.29 |
| P value | 0.044 | 0.023 | 0.016 | 0.010 | 0.032 | 0.000 | 0.001 |

Finally, our findings are consistent with those reported in the literature. This can be a useful step towards obtaining reference values or creating databases useful for the scientific community to foster the clinical utility of HRV analysis.

## 5 Conclusion

HRV values are instrumental in identifying cardiovascular risk. Therefore, establishing reference values for the healthy population could aid in the early detection and prevention of cardiac diseases.

A review of the literature reveals inconsistent HRV parameter values in healthy subjects across various studies. Consequently, the lack of standardized HRV analysis benchmarks complicates the comparison of research works and the interpretation of specific values obtained under different conditions. Additionally, HRV indices can be analysed across diverse domains: time, frequency, and nonlinear.

In previous studies, some preliminary results of HRV indices, offering guidelines for planning HRV studies, were published. However, a major limitation of those works was the small size of the database used. In this work, we sought to validate our previous findings by significantly expanding the database size while maintaining consistent acquisition and analysis protocols. We conducted a comprehensive analysis of the three types of HRV indices, focusing on their normal distributions, stability, and parameter correlations. Additionally, a reproducibility study using the public Fantasia Database was undertaken. Finally, we also compared our results with other HRV data from healthy subjects in similar studies.

Our findings indicate that not all HRV parameters are equally effective, with some being correlated and therefore redundant. We recommend ApEn as a suitable index, and IRRR may be preferable over SDNN for quantifying global variability. For LF-related parameters, HRVi and SD2 are advisable, and for HF-related indices, either MADRR or rMSSD could be chosen.

While further research with more extensive cases is still necessary, we believe our work contributes valuable baseline values. This can assist clinicians and researchers in standardizing HRV studies under normality conditions.